%% file: acl2021.tex
\documentclass[11pt,a4paper]{article}
\usepackage[hyperref]{acl2021}
\usepackage{times}
\usepackage{latexsym}

\usepackage{microtype}

\usepackage{graphicx}
\usepackage{booktabs,multirow}
\usepackage{caption}
\usepackage[whole]{bxcjkjatype}
\usepackage{bbding}
\usepackage{pifont}
\usepackage{subcaption}
\usepackage{lingmacros}
\usepackage{tikz-dependency}
\usepackage{pifont}
\usepackage{amssymb,mathtools}

\usepackage{color}
\usepackage{caption}
\usetikzlibrary{positioning}  

\newcommand{\gl}[2]{%
\leavevmode\vtop{\hbox{#1}%
\hbox{#2\lower1.4ex\rlap{ }}}}
\newcommand\doublecheck{\checkmark\kern-0.4em\checkmark}

\aclfinalcopy %
\def\aclpaperid{2755} %

\title{Lower Perplexity is Not Always Human-Like}

\author{Tatsuki Kuribayashi$^{1,2}$, Yohei Oseki$^{3,4}$, Takumi Ito$^{1,2}$, \\
{\bf Ryo Yoshida$^{3}$, Masayuki Asahara$^{5}$, Kentaro Inui$^{1,4}$} \\
 $^1$Tohoku University 
 $^2$Langsmith Inc. 
 $^3$University of Tokyo
 $^4$RIKEN 
 $^5$NINJAL \\
\texttt{\{kuribayashi, takumi.ito.c4, inui\}@tohoku.ac.jp }, \\
\texttt{\{oseki, yoshiryo0617\}@g.ecc.u-tokyo.ac.jp }, 
\texttt{masayu-a@ninjal.ac.jp}}

\date{}

\begin{document}
\maketitle

\begin{abstract}
In computational psycholinguistics, various language models have been evaluated against human reading behavior (e.g., eye movement) to build human-like computational models. However, most previous efforts have focused almost exclusively on English, despite the recent trend towards linguistic universal within the general community. In order to fill the gap, this paper investigates whether the established results in computational psycholinguistics can be generalized across languages. Specifically, we re-examine an established generalization ---\textit{the lower perplexity a language model has, the more human-like the language model is}--- in Japanese with typologically different structures from English. Our experiments demonstrate that this established generalization exhibits a surprising lack of universality; namely, lower perplexity is not always human-like. Moreover, this discrepancy between English and Japanese is further explored from the perspective of (non-)uniform information density. Overall, our results suggest that a cross-lingual evaluation will be necessary to construct human-like computational models.
\end{abstract}

\section{Introduction}
\label{sec:intro}
It is well known that the probability of a word in context (i.e., surprisal) impacts its processing difficulty in incremental human language comprehension~\cite{hale-2001-probabilistic,Demberg2008DataComplexity,Levy2008Expectation-basedComprehension,Smith2013TheLogarithmic}.
Building on this basis, researchers have compared a variety of language models (LMs) in terms of how well their surprisal correlates with human reading behavior~\cite{roark-etal-2009-deriving,frank2011insensitivity,fossum-levy-2012-sequential,Hale2018FindingSearch,Goodkind2018PredictiveQuality,aurnhammer2019comparing,Merkx2020ComparingData,Wilcox2020OnBehavior}. 
Such investigations could provide insights into the development of a general computational model of human language processing.
For example, recent studies reported that LMs with better performance for next-word prediction could also better predict the human reading behavior (i.e. more human-like)~\cite{fossum-levy-2012-sequential,Goodkind2018PredictiveQuality,Wilcox2020OnBehavior}.

{\setlength\textfloatsep{0pt}
\begin{figure*}[t]
    \centering
      \includegraphics[width=13.5cm]{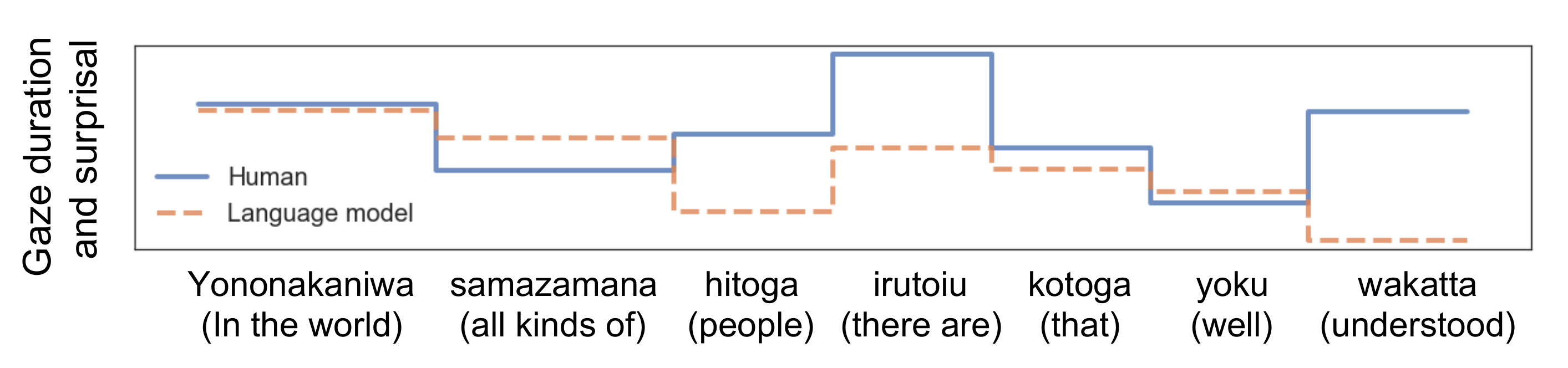}
      \caption{Gaze duration from human subjects and surprisal from language models for the Japanese sentence ``Yononakaniwa samazamana hitoga irutoiu kotoga yoku wakatta.'' (\textit{I understood well that there are all kinds of people in the world.})}
      \label{fig:intro}
\end{figure*}
}

In this paper, we re-examine whether the recent findings on human-like computational models can be generalized across languages.
Despite the community's ongoing search for a language-independent model~\cite{bender2011achieving}, existing studies have focused almost exclusively on the English language.
Having said that, broad-coverage cross-linguistic evaluation of the existing reports is prohibitively difficult.
In fact, data on human reading behavior (e.g., eye movement) is available only in limited languages.
As an initial foray, this study focuses on the Japanese language as a representative of languages that have typologically different characteristics from the English language.
If the observation is different between English and Japanese, the current findings on English data might lack a universality across languages.

We specifically revisit the recent report---\textit{the lower perplexity a LM has, the more human-like the LM is}---in the English and Japanese languages~\cite{fossum-levy-2012-sequential,Goodkind2018PredictiveQuality,Wilcox2020OnBehavior}.
In addition to the importance of cross-linguistic evaluation, the report itself is worth investigating.
Recent studies in the machine learning field have reported that more parameters, training data, and computation cost can result in better PPL~\cite{kaplan2020scaling,DBLP:conf/nips/BrownMRSKDNSSAA20}.
Our investigation has implications for whether a human-like model might exist beyond such improvements.

More concretely, over three dozens of LMs were trained for each language, with variants in their architecture, training data size, and the number of parameter updates. 
Then, the surprisals computed by each LM were compared to human eye movement data (Figure~\ref{fig:intro}).
The analysis of the relationship between PPL and the psychometric predictive power revealed substantively different trends between the Japanese and English LMs. 
In Japanese, a lower PPL of a LM does not indicate better performance for modeling reading behavior.
By contrast, in English, there was a clear relationship between the two metrics as reported in the prior studies.

This opens a remaining and important question: why are English and Japanese different in this aspect? 
We discuss the differing results between English and Japanese from the perspective of the uniform information density hypothesis~\cite{genzel-charniak-2002-entropy,levy2005probabilistic,NIPS2006_c6a01432}.
We find that the processing difficulty (i.e., gaze duration) of segments is less uniformly distributed within a Japanese sentence.
Given this, the discrepancy of the results between English and Japanese might stem from a mismatch between the information uniformity of the target language and the LM's training objective.
We demonstrate that tuning Japanese LMs to this training objective collapses the human-like nonuniformity of the processing difficulty observed in Japanese subjects.
Our code is made publicly available.\footnote{\url{https://github.com/kuribayashi4/surprisal_reading_time_en_ja}}

\section{Related work}
\label{sec:rel}

\subsection{Human sentence processing and LMs}
\label{subsec:human}

What factor determines the incremental difficulty of human language processing?
At present, surprisal theory~\cite{hale-2001-probabilistic,Levy2008Expectation-basedComprehension} has been widely adopted in the field of computational psycholinguistics.
This theory suggests that the processing difficulty of a segment is determined by how predictable the segment is in its preceding context ($-\log{p(\mathrm{segment}| \mathrm{preceding\ context})}$).

Existing studies have compared various computational models by checking the effectiveness of their surprisals in modeling human reading behavior~\cite{hale-2001-probabilistic,roark-etal-2009-deriving,frank2011insensitivity,fossum-levy-2012-sequential,Hale2018FindingSearch,Goodkind2018PredictiveQuality,Merkx2020ComparingData,Wilcox2020OnBehavior}.
Data such as eye movement~\cite{kennedy2003dundee} and brain activity~\cite{FRANK20151,brennan2016abstract} are used as measures of human reading behavior.
For example, using eye movement data, \citet{frank2011insensitivity} compared the surprisals from phrase-structure grammars (PSGs) with those from a non-hierarchical, sequential model, tentatively concluding that human sentence processing was insensitive to hierarchical structures since non-hierarchical models displayed better psychological predictive power than PSGs.
Recently, researchers reported that surprisals from LMs with low PPL correlate well with human reading behaviors~\cite{fossum-levy-2012-sequential,Goodkind2018PredictiveQuality,aurnhammer2019comparing,Wilcox2020OnBehavior}.

The work most closely related to this study is~\citet{Wilcox2020OnBehavior}.
They examined the relationship between PPL, psychometric predictive power, and syntactic knowledge in LMs using a variety of models, including modern neural LMs~\cite{Radrof2018LanguageLearners}.
They found a tight relationship between PPL and psychometric predictive power in the English corpora. 
This study investigates whether this relationship can be generalized across languages.

\subsection{Reading behavior in Japanese}
\label{subsec:eng_ja}

In comparison to English speakers, Japanese speakers display different patterns in sentence processing.  
For example, an anti-locality effect (the more modifiers a word has in its preceding context, the easier the word is to process) has typically been observed in head-final languages, including Japanese~\cite{Konieczny2000LocalityComplexity}.
Such differences between the languages are assumed to be more or less due to their different sentence structures.
Recently, eye movement data for naturally occurring Japanese texts have recently become available~\cite{Asahara2016Reading-TimeJapanese} and was extensively annotated with various linguistic properties~\cite{Asahara2017BetweenCategories,Asahara2017BetweenStructure,Asahara2018BetweenLanguage}.

\section{Methods}
\label{sec:method}

This section describes the settings of LMs, eye movement data, and evaluation metrics.

\subsection{Language models}
A variety of sentence-level, left-to-right sequential LMs was used.

\paragraph{Training data of English LMs:}
We used the WikiText-103 dataset to train the English LMs.
Based on the reports that subword-level English LMs exhibits superior psychometric predictive power~\cite{Wilcox2020OnBehavior}, input texts were divided into subwords by a byte-pair encoding (BPE)~\cite{Sennrich2016NeuralUnits}.\footnote{Implemented in SentencePiece~\cite{Kudo2018SentencePiece:Processing}. We set character coverage to 0.9995，and vocabulary size to 32,000 in English. In Japanese, the vocabulary size is 100,000, reflecting its rich morphemes.}
The training data consist of approximately 4M sentences (114M subwords units).

\paragraph{Training data of Japanese LMs:}
We used news articles and the Japanese part of Wikipedia to train the Japanese LMs.
Input texts were first segmented into morphemes by MeCab~\cite{kudo2006mecab} with unidic dictionary, and then further divided into subwords by BPE.\footnotemark[2]
The training data consist of approximately 5M sentences (146M subwords units).

\paragraph{Architectures:}
The following four variants of LMs were used: Transformer-large (\textsc{Trans-lg})~\cite{Vaswani2017AttentionNeed}, Transformer-small (\textsc{Trans-sm}), LSTM (\textsc{LSTM})~\cite{Hochreiter1997LongMemory}, and N-gram LMs (\textsc{N-gram}).\footnote{The neural LMs were trained with the fairseq toolkit~\cite{ott-etal-2019-fairseq}. \textsc{N-gram} LMs were trained using KenLM \url{https://github.com/kpu/kenlm}.}
The parameter size was almost the same for \textsc{Trans-sm} and \textsc{LSTM}.
With respect to the \textsc{N-gram} models, 3-gram, 4-gram, and 5-gram LMs were used.
Appendix~\ref{app:hyper} shows the hyperparameters of the neural LMs.

\paragraph{Training data size:}
For each neural LM architecture (\textsc{Trans-lg}, \textsc{Trans-sm}, and \textsc{LSTM}), three variants were trained using different training data sizes: \textsc{lg} (full training data), \textsc{md} (1/10 training data), and \textsc{sm} (1/100 training data).
The N-gram LMs were trained on \textsc{lg} datasets.

\paragraph{Number of updates:}
The parameters of each neural LM were saved at four different points during training: 100, 1K, 10K, and 100K parameter updates.

To summarize, 39 LM training settings were attained for each language (3 architectures $\times$ 3 data size $\times$ 4 parameter updates $=$ 36 neural LMs, plus 3 \textsc{N-gram} LMs).
In addition, our experiments use three LMs trained using different random seeds for each neural LM training configure; hence, 111 LMs (36 neural LMs $\times$ 3 seeds, plus 3 \textsc{N-gram} LMs) were tested for each language.

\begin{table*}[t]
\centering
\renewcommand{\arraystretch}{0.8}
{\small
\begin{tabular}{lrrrp{1.5cm}p{1.5cm}p{1.5cm}p{2.2cm}} \toprule
Corpus & \#articles &  \#sents. & \#segments & \#data points (used) & \#subjects per article & Avg. GD per segment &  Avg. \#subwords per segment \\
\cmidrule(lr){1-1} \cmidrule(lr){2-2} \cmidrule(lr){3-3} \cmidrule(lr){4-4} \cmidrule(lr){5-5} \cmidrule(lr){6-6} \cmidrule(lr){7-7} \cmidrule(lr){8-8}
Dundee Corpus & 20 & 2,478 & 51,501 & \multicolumn{1}{r}{214,955}  & \multicolumn{1}{r}{10} & \multicolumn{1}{r}{227.1} & \multicolumn{1}{r}{1.3} \\
BCCWJ-EyeTrack & 20 & 218 & 1,643 & \multicolumn{1}{r}{6,009} & \multicolumn{1}{r}{12} & \multicolumn{1}{r}{361.6} & \multicolumn{1}{r}{3.4} \\
\bottomrule
\end{tabular}
}
\caption{Statistics of the corpora used for evaluating the psychometric predictive power of LMs. ``\#articles'' and  ``\#sents.'' are the number of articles and sentences in each corpus. ``\#segments'' denotes the number of segments annotated with human reading time in each corpus. ``\#data points'' is the number of reading time annotations used in our experiments. Each segment has the reading time annotations from multiple subjects (\#subjects per article). ``Avg. GD per segment'' is the averaged gaze duration per segment. ``Avg. \#subwords per segment'' denotes the averaged number of subwords consisting of each segment.
}
\label{tbl:stats}
\end{table*}

\subsection{Eye movement data}
\label{subsec:eye_track_data}

\paragraph{English:} The Dundee Corpus~\cite{kennedy2003dundee}, which contains gaze duration annotation for each word, was used.
Following~\citet{Smith2013TheLogarithmic}, the first-pass gaze duration was analyzed.
Then, following~\citet{Goodkind2018PredictiveQuality}, the data points that met any of the following criteria were excluded:
\begin{itemize}
\setlength{\parskip}{0cm} 
\setlength{\itemsep}{0.1cm}
    \item data points with zero gaze duration or that beyond three standard deviations
    \item segments with punctuation or numeric characters
    \item segments whose next segment has punctuation or numeric characters 
    \item first or last segment in a line
\end{itemize}
In total, the analysis included 214,955 data points in the corpus.

\paragraph{Japanese:}
The BCCWJ-EyeTrack~\cite{Asahara2016Reading-TimeJapanese}, which contains gaze duration annotation for each phrasal unit, was used.
Note that the phrasal unit (i.e., bunsetsu) consists of at least one content morpheme and its postpositional function morphemes.
Henceforth, an English word and a Japanese phrasal unit are referred to as a ``segment.''
The same exclusion criteria as the Dundee Corpus was applied to the BCCWJ-EyeTrack data.\footnote{Strictly speaking, the exclusion criteria was slightly different between Japanese and English data. In the Japanese data, we included the segments whose next segment had punctuation or a numeric character, as there is no spillover effect in Japanese (see Section~\ref{subsec:metric})}
In total, the analysis included 6,009 data points in the corpus.
Note that the BCCWJ-EyeTrack data was deliberately designed to address language-specific issues in Japanese such as the lack of segmentation spaces in Japanese texts~\cite{Asahara2016Reading-TimeJapanese}.

\paragraph{Statistics:}
Table~\ref{tbl:stats} shows the statistics of the Dundee Corpus and BCCWJ-EyeTrack data.
The BCCWJ-EyeTrack has more than 10 times a smaller number of data points than the Dundee Corpus.
Notably, the segment annotated with eye movement information differs between English and Japanese.
On average, a Japanese segment consists of 3.4 subwords, while an English segment consists of 1.3 subwords.
\citet{Smith2013TheLogarithmic} theoretically proved that the more fragments a word is divided into when computing its surprisal, the better the calculated surprisal approximates the human cognitive effort if the human language processing is highly incremental.
Thus, we tentatively consider that this difference did not make a negative impact on the results using the Japanese data.

\subsection{Evaluation metrics}
\label{subsec:metric}

\paragraph{Perplexity (PPL):}
\label{subsubsec:perplexity}

PPL, the inverse geometric mean of next-word probabilities $p(w_{i}|w_{<i})$ in a text that consists of $N$ signals ($w_1, w_2, \cdots, w_N$), is a typical evaluation metric for unidirectional LMs (Eq.~\ref{eq:ppl}):

\begin{align}
    \text{PPL} &=\prod_{i=0}^N p(w_{i}|w_{<i})^{-\frac{1}{N}} \:\:.
    \label{eq:ppl}
\end{align}

{\setlength\textfloatsep{0pt}
\begin{figure*}[t]
    \centering
      \includegraphics[width=\linewidth]{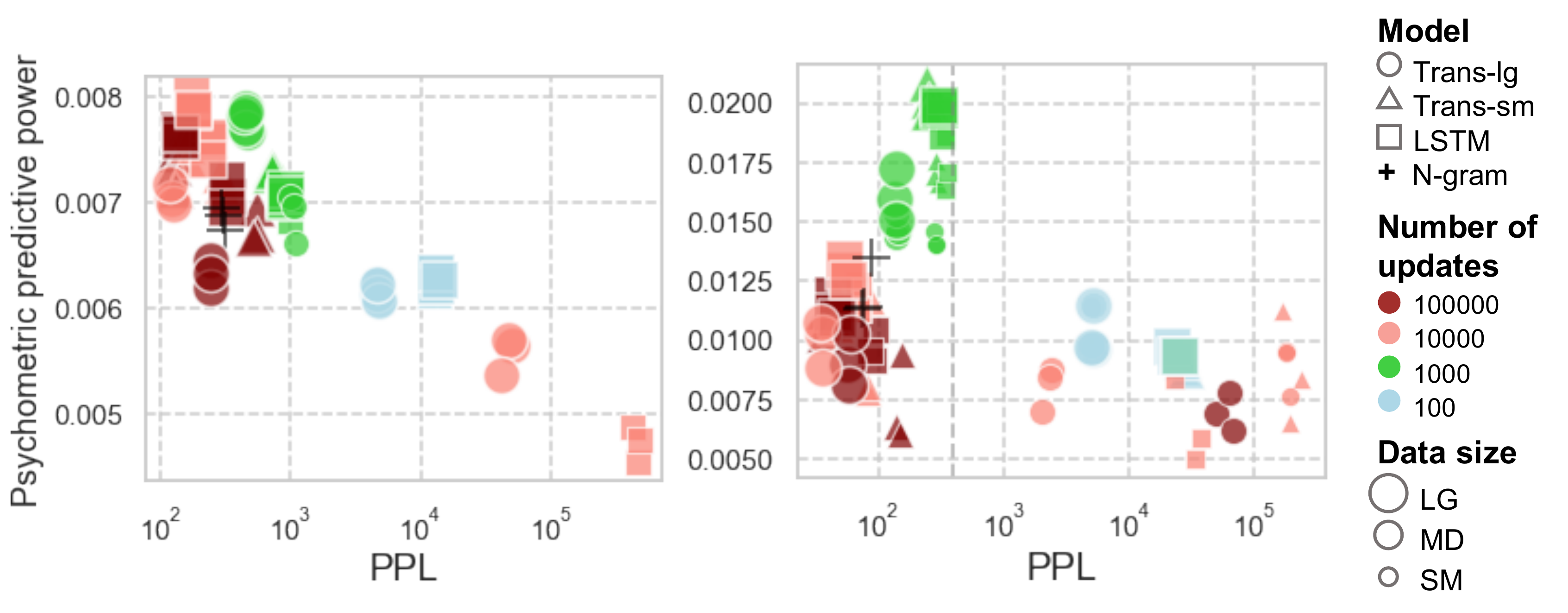}
      \caption{Relationship between PPL (X-axis) and psychometric predictive power, i.e., $\Delta$LogLik (Y-axis) in the English and Japanese languages. Each point corresponds to each LM. A low score on the X-axis indicates the high linguistic accuracy of the model. The PPL was calculated on the eye movement data, and the LMs with PPL more than 10$^6$ were excluded from the figure. A high score on the Y-axis indicates that the model has a high psychometric predictive power. Note that the X-axis is on a log scale.}
      \label{fig:pred_power}
\end{figure*}
}
\noindent
Low PPL indicates that the model can accurately predict the upcoming signal based on its preceding context.
The training objective of LMs works to minimize the PPL computed by the model.
In the experiments, the PPL of a LM is evaluated with the texts in the eye movement data, which do not overlap with the training data.
A model with low PPL is also called a \textit{linguistically accurate} model~\cite{frank2011insensitivity}.

\paragraph{Psychometric predictive power:}
The surprisal measure, a negative logarithmic probability of a segment in context ($-\log{p({\rm segment}| {\rm preceding\ context})}$), is a widely used information-theoretic complexity metric.
Intuitively, a model is considered to have high psychometric predictive power (i.e., \textit{psychological accuracy}) if the surprisals of segments computed by the model have trends similar to the human subject's cognitive load (e.g., measured by gaze duration).
Following the existing studies~\cite{Goodkind2018PredictiveQuality,Merkx2020ComparingData,Wilcox2020OnBehavior}, the psychometric predictive power of a model was measured by comparing surprisal from the model and gaze duration from human subjects.

While LMs process a text subword-by-subword, gaze duration is annotated in a larger segment.
Following the study using subwords~\cite{Wilcox2020OnBehavior}, the surprisal of each segment was calculated using the joint probability of its constituent subwords.
Formally, given a text consisting of $N$ subwords $w_{1:N}=(w_1,w_2,\cdots,w_N)$, surprisal $I(\cdot)$ of a segment $s_k = (w_l,w_{l+1},\cdots,w_{m})$, where $1\leq l \leq m \leq N$, was calculated as follows:

\begin{align}
\begin{split}
    I(s_k) &= -\log p(w_l,\cdots,w_m|w_{< l}) \\
    &= -\sum_{k=l}^{m}\log p(w_k|w_1,\cdots,w_{k-1})  \:\:.
\end{split}
    \end{align}

The effect of surprisals for modeling human reading behavior was calculated using a linear mixed-effects regression~\cite{JSSv067i01}.
Specifically, the gaze duration (\texttt{GD}) was modeled using the following formula: 

{\small
\begin{align}
  \begin{split}
    \texttt{GD} &\sim \texttt{surprisal} + \texttt{surprisal\_prev\_1} \\
    &+ \texttt{surprisal\_prev\_2} + \texttt{freq}*\texttt{length} \\
    &+ \texttt{freq\_prev\_1}*\texttt{length\_prev\_1} \\
    &+ \texttt{screenN} + \texttt{lineN} + \texttt{segmentN} \\
    &+ \texttt{(1|article)} + \texttt{(1|subj)} \:\:.
    \label{eq:regression}
  \end{split}
\end{align}
}

\noindent
The regression model includes baseline factors (e.g., frequency of a segment) that are of no interest in the comparison of LMs.
A collection of factors used in the existing studies~\cite{Asahara2016Reading-TimeJapanese,Wilcox2020OnBehavior} were initially examined and the factors that were not significant ($p>0.05$) for gaze duration modeling both in the Dundee Corpus and BCCWJ-EyeTrack were excluded.
The frequency of a segment (\texttt{freq}) was calculated using the entire training data for LMs.
Appendix~\ref{app:feature} shows the details of each factor in Eq.~\ref{eq:regression}.

In English experiments, surprisals of preceding words (\texttt{surprisal\_prev\_1} and \texttt{surprisal\_prev\_2}) were included in order to handle the spillover effect (the processing cost of a certain segment is affected by its preceding segments)~\cite{rayner1996effects,Smith2013TheLogarithmic}.
In Japanese experiments, the surprisals of preceding words were not included because our preliminary experiment showed that these factors were not significantly effective for modeling gaze duration in the BCCWJ-EyeTrack.\footnote{The reason is probably that a Japanese phrasal unit (i.e., bunsetsu) could be a larger unit than an English word.}
All the regression models used in our experiments were converged.

{\setlength\textfloatsep{0pt}
\begin{figure*}[t]
    \centering
      \includegraphics[width=15cm]{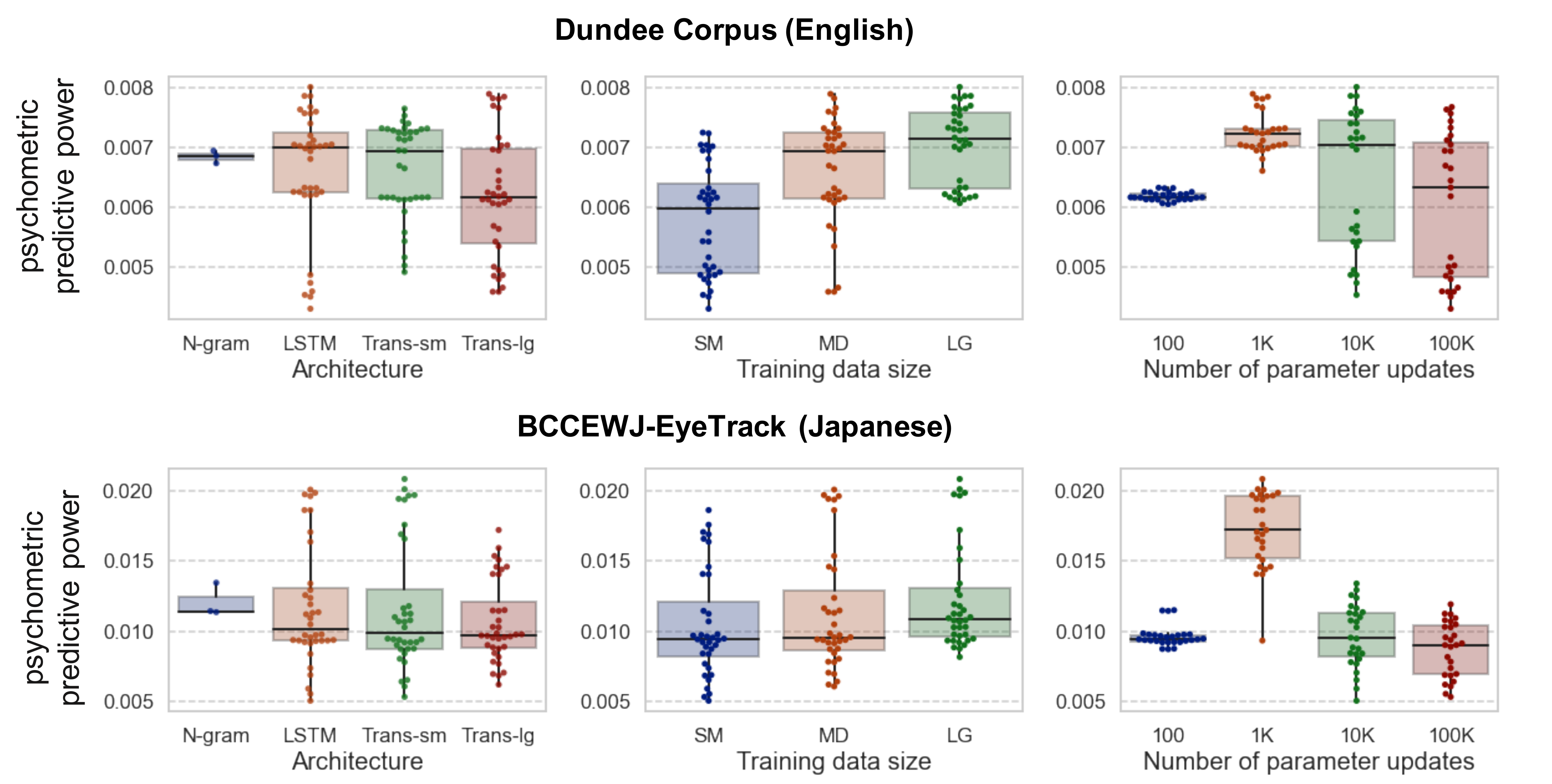}
      \caption{Separate effect of model architecture, training data size, and the number of parameter updates for LMs' psychometric predictive power in each language. Each point corresponds to each LM. The box shows the quartiles of the data. The whiskers show 1.5 times interquartile range.}
      \label{fig:ana}
\end{figure*}
}

To isolate the effect of surprisal for gaze duration modeling, a baseline regression model was trained without surprisal information (excluding the \texttt{surprisal}, \texttt{surprisal\_prev\_1}, and \texttt{surprisal\_prev\_2} terms from Eq.~\ref{eq:regression}).
Following~\citet{Wilcox2020OnBehavior}, the mean by-segment difference of log-likelihood between the model using surprisal values (Eq.~\ref{eq:regression}) and the baseline model was calculated.
Henceforth, this metric is called $\Delta$LogLik.
When surprisal from a LM is not effective for gaze duration modeling, the $\Delta$LogLik score becomes zero.
A high $\Delta$LogLik means that the surprisal values obtained by the LM are effective for modeling gaze duration (i.e., the LM has a high psychometric predictive power).

\section{Experiments}
\label{sec:exp1}

The relationship between PPL and psychometric predictive power is investigated.
Furthermore, the relationship is analyzed with respect to the training configures of LMs (e.g., the number of parameter updates).
Then, we discuss the results from the perspective of the uniformity of information density.

\subsection{Psychometric predictive power and PPL}
\label{subsec:result_1}

Figure~\ref{fig:pred_power} shows the relationship between PPL and psychometric predictive power (i.e., $\Delta$LogLik) of LMs in each of the languages.
Each point corresponds to each LM, and a score on the Y-axis indicates the psychometric predictive power of a LM (higher is better).
The X-axis is PPL on a log scale (lower is better).

\paragraph{Dundee Corpus:}
First, the results of the data from the Dundee Corpus show a clear relationship between PPL and psychometric predictive power; namely, lower PPL corresponds to more psychometric predictive power, as reported by prior studies~\cite{Goodkind2018PredictiveQuality,Wilcox2020OnBehavior}.
Spearman's rank correlation coefficient between the two metrics was $-0.87$.

\paragraph{BCCWJ-EyeTrack:}
By contrast, in BCCWJ-EyeTrack, there was no clear, consistent trend between the PPL and psychometric predictive power.
While LMs with PPL over 400 show the correlation between PPL and psychometric predictive power ($-0.68$ with Spearman's $\rho$), there is a positive correlation ($0.53$ with Spearman's $\rho$) for LMs with PPL below 400.
The positive correlation means that the more accurately the LMs can predict the upcoming word, the \textit{worse} the psychometric predictive power of the LMs is. 
These results demonstrate the non-universality of the recent report across languages; \textit{lower perplexity is not always human-like}.
The \textsc{lstm} LM trained using the \textsc{md} dataset with 1K updates achieved the best psychometric predictive power.
Notably, surprisal was effective for gaze duration modeling in all the Japanese LMs.
$\Delta$logLik scores were significantly higher than zero with the chi-square test ($p<$0.05).

\subsection{Model architectures, data sizes, number of parameter updates}
\label{subsec:result_2}

Which factor (e.g., model architecture, training data size, and the number of parameter updates) characterizes the psychometric predictive power of LMs?
Is the collection of effective factors consistent between the two languages?
This study takes a more in-depth look at the separate effects of (i) model architecture, (ii) training data size, and (iii) the number of parameter updates for the psychometric predictive power.

Figure~\ref{fig:ana} summarizes the effect of each factor, where the Y-axis denotes the psychometric predictive power.
The most noticeable trend is that Japanese LMs with a relatively fewer number of parameter updates (1K) have better psychometric predictive power than the other Japanese LMs (bottom right part of Figure~\ref{fig:ana}), while this trend does not exist in the English LMs (top right part).
This implies that the training objective of the LMs, maximizing $\frac{1}{N}\sum_{i=1}^{N}\log P(w_i|w_{<i})$, had a negative impact on the psychometric predictive power of LMs, at least in Japanese.
We discuss this point in Section~\ref{subsec:discussion}.

To quantitatively test the differences in Figure~\ref{fig:ana}, a linear regression model was trained to estimate psychometric predictive power with the factors of the model architecture, the training data size, and the parameter update number in each language.
The training data size and the parameter update number are represented as logarithmically transformed numerical factors.
The following trends were found: (i)  ; (ii) the training data size positively affects the performance in English alone; and (iii) the number of parameter updates positively affects the performance only in English.
There was no factor that boosted the psychometric predictive power of LMs in both English and Japanese languages.

\subsection{Discussion: uniform information density}
\label{subsec:discussion}
The key question is: why do Japanese and English show different trends between PPL and psychometric predictive power?
One possible interpretation connecting our results to the uniform information density is discussed in this section.

In computational psycholinguistics, it is commonly assumed that language is designed to enable efficient communication.
This principle has been typically investigated under the uniform information density (UID) hypothesis~\cite{genzel-charniak-2002-entropy,levy2005probabilistic,NIPS2006_c6a01432}.
This hypothesis suggests that speakers seek to keep the amount of information constant across the signals (e.g., segments).

Assuming this hypothesis holds for all languages, the reasonable expectation would be for human subjects to show a near-uniform gaze duration across segments regardless of their native language.
However, this study found that the coefficient of variation\footnote{Coefficient of variation is $\frac{\sigma}{\mu}$, where $\sigma$ and $\mu$ are the standard deviation and the mean of the first-pass gaze durations in the eye movement data.} in gaze duration over the whole corpus was around 1.7 times higher in Japanese compared to English (0.75 vs. 0.44).
Specifically, in Japanese, the gaze duration tended to speed up towards the end of sentences, whereas the duration was near-uniform in English (Figure~\ref{fig:uniformity}).\footnote{At least in our experimental setup, token position within the sentence was not significantly effective for gaze duration modeling in English sentences, whereas it was significant in Japanese sentences. We checked the coefficient of the factor of position in sentence \texttt{segmentN} using the linear regression model of \texttt{GD $\sim$ sengmentN}.}
These observations imply that the Japanese language might have a less uniform information density than English.
This phenomenon was also investigated through the lens of word order, where SOV languages such as Japanese are reported to show less uniformity of information density~\cite{NIPS2010_0c74b7f7}. 

Based on this observation, the discrepancy between English and Japanese low-PPL LMs' psycholinguistic predictive power could stem from a mismatch between the LM's training objective and the information uniformity of the target language. 
The objective function, $\frac{1}{N}\sum_{i=1}^{N}\log P(w_i|w_{<i})$, defines that the ``ideal'' is to maximize all next word probabilities to 1.0 (a \textit{uniform} goal).\footnote{PPL, $\prod_{i=1}^{N}P(w_i|w_{<i})^{-\frac{1}{N}}$, is minimized when the LM objective are maximized.} 
That is, LMs are, \textit{in theory}, trained to approach a model satisfying the UID assumption~\cite{bloem2016testing}, where all surprisals from the LM are equally, sufficiently small across the segments.
Therefore, the objective function might lead to a worse approximation of human-like surprisal in languages that are further from the UID assumption, such as Japanese, while it might be more compatible with English, which has a more uniform processing difficulty across segments.
This explanation would be consistent with the observation that more tuning to the LM training objective (i.e., a lower PPL) had a negative impact on the psycholinguistic performance of the Japanese LMs (Section~\ref{subsec:result_2}).
Note the tendency of LMs to assign unreasonably high probabilities to segments has also attracted attention from the viewpoint of memorization capability of LMs~\cite{DBLP:journals/corr/abs-2012-07805}.
In addition, the connection of the UID hypothesis to the modern NLP techniques has been recently explored~\cite{meister-etal-2020-beam,wei2021cognitive}. 
We further investigate our hypothesis in Section~\ref{sec:exp2}.

{\setlength\textfloatsep{0pt}
\begin{figure}[t]
    \centering
      \includegraphics[width=7.5cm]{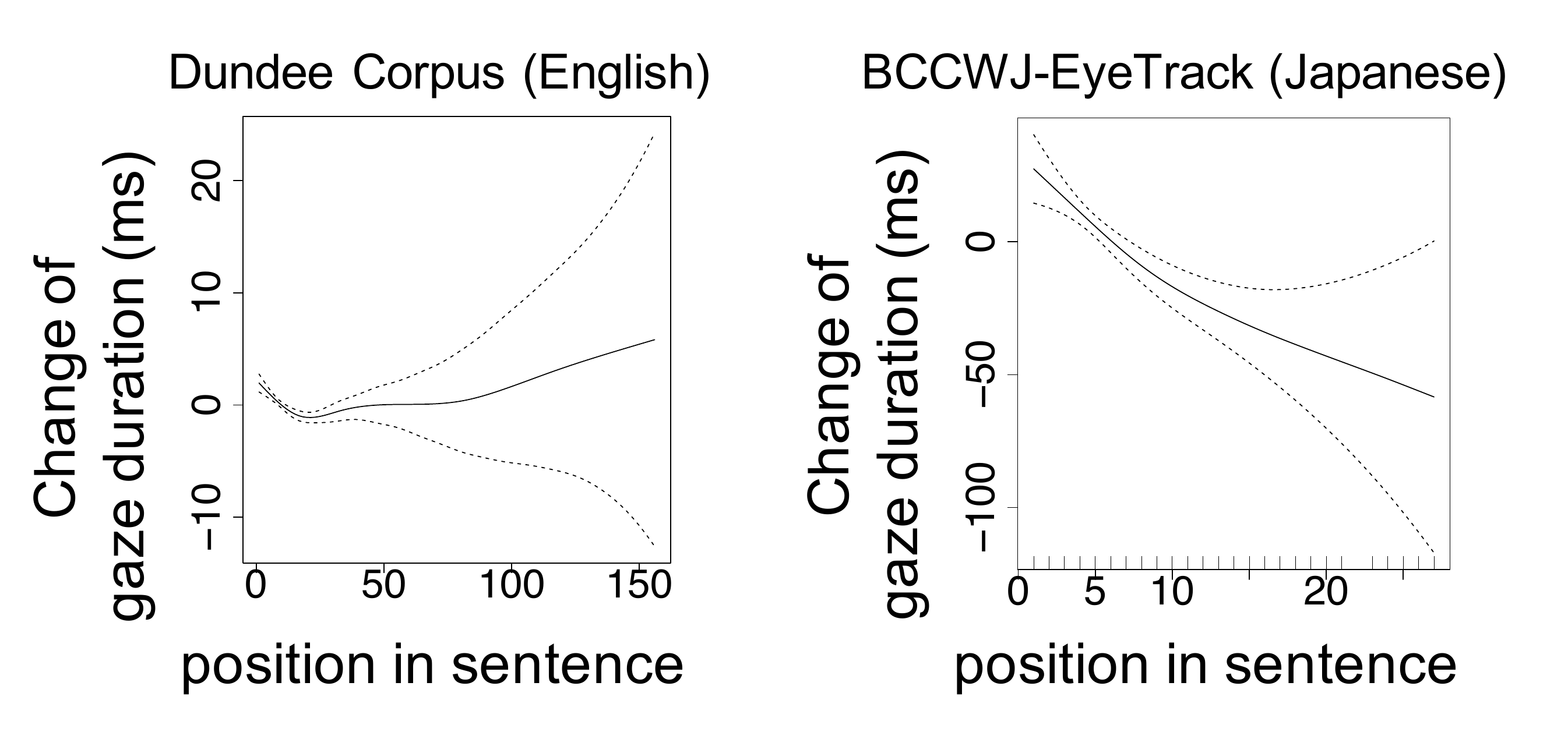}
      \caption{Uniformity of gaze duration with respect to segment position in a sentence. This plot is computed by the generalized additive model of \texttt{GD $\sim$ segmentN}. Here, \texttt{segmentN} is denoted as the position of a segment in a sentence.}
      \label{fig:uniformity}
\end{figure}
}

{\setlength\textfloatsep{0pt}
\begin{figure*}[t]
    \centering
      \includegraphics[width=\hsize]{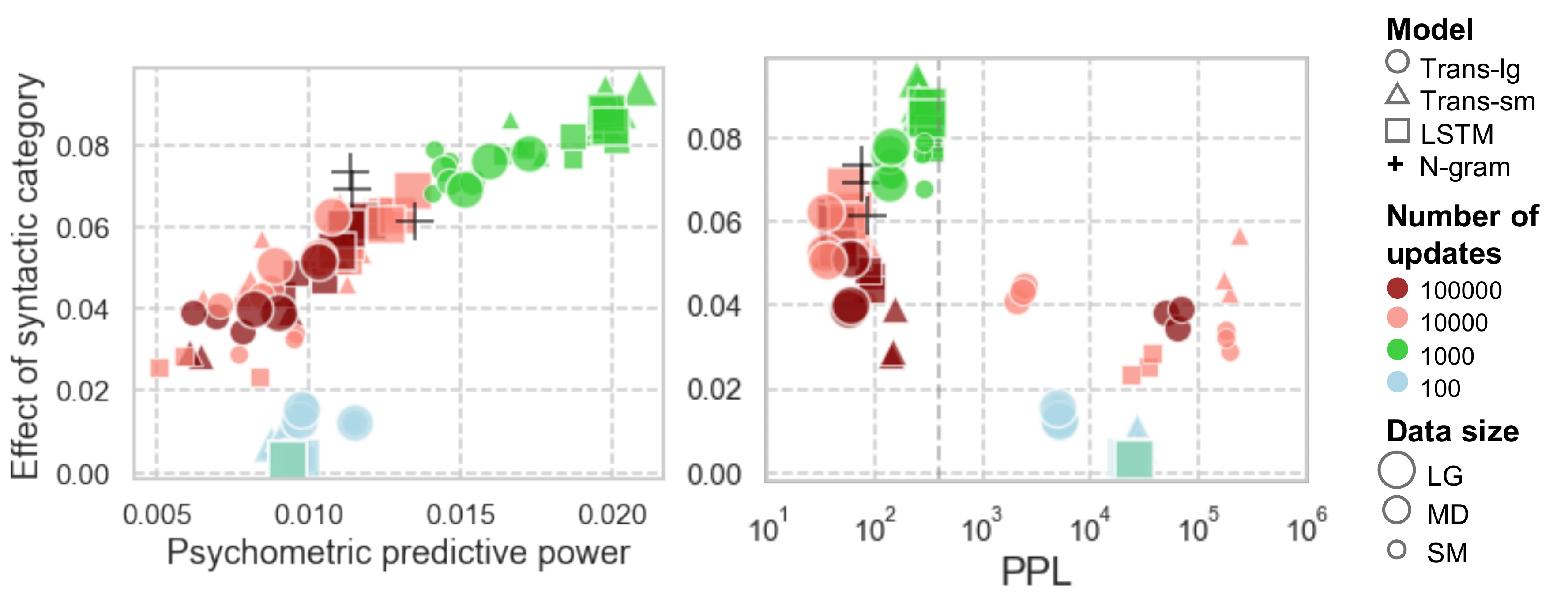}
      \caption{Relationship between the LM's psychometric predictive power and the effect of the syntactic category on the surprisal computed by each LM (left part), and that between PPL and the effect of the syntactic category (right part). Each point corresponds to each LM. The PPL was calculated on the eye movement data, and the LMs with PPL more than 10$^6$ was excluded from the right part of the figure.}
      \label{fig:syn_ana}
\end{figure*}
}

\section{Probing nonuniform information density of Japanese LMs}
\label{sec:exp2}

This study hypothesized that tuning to the LM objective (i.e., uniform goal) obscures the nonuniform trend observed in the reading behavior of Japanese subjects.
We investigated whether the nonuniformity of the processing difficulty observed in human reading time is mirrored by LM surprisals.

\paragraph{Settings:}
In a preliminary experiment, we observed that the \emph{syntactic category} (similar to part-of-speech) was the most dominant linguistic factor for explaining the difference in human gaze duration in Japanese sentences (see Appendix~\ref{app:pre_exp2}).
Based on this observation, we analyze the nonuniformity of surprisals in Japanese LMs with respect to the syntactic categories.

The segments in BCCWJ-EyeTrack were classified into one of the following syntactic categories: (a) \texttt{nominal} (nouns), (b) \texttt{verbal} (verbs), (c) \texttt{modifier} (adjectives and adverbs), and (d) \texttt{other} entries, as follows: 

\begin{center}
  \begin{dependency}[theme=simple]
    \begin{deptext}[column sep=0.1em]
      \textit{Kanojo-ga} \& \textit{akai} \& \textit{kaban-o} \& \textit{kat-ta} \\
      She-\texttt{NOM} \& red \& bag-\texttt{ACC} \& buy-\texttt{PAST} \\
      \texttt{nominal} \& \texttt{modifier} \& \texttt{nominal} \& \texttt{verbal} \\
    \end{deptext}
  \end{dependency}
\end{center}

\noindent
As~\citet{Asahara2017BetweenCategories} reported, \texttt{verbal} and \texttt{modifier} segments have a shorter gaze duration than the other segments in Japanese sentences.
An analysis was conducted on how strongly the Japanese LM's surprisals on segments are influenced by their syntactic category.
This influence can be evaluated by examining how effectively syntactic category factors can model LM surprisals.

In this experiment, surprisal was regarded as ``simulated gaze duration'' from an ``LM subject,'' and the importance of syntactic category information for modeling the simulated gaze duration (\texttt{simulated\_GD}) was evaluated.
To inspect the effect of the syntactic category information for modeling the simulated gaze duration, the following regression model\footnote{\texttt{sentN} and \texttt{tokenN} denote the sentence position and the segment position in a sentence (see Appendix~\ref{app:feature}). Note that the \texttt{tokenN} and syntactic category exhibit low correlation (0.02 with Pearson's $r$).} was used, including a factor defining which syntactic category the segment falls into (\texttt{syn\_category}):

{\small
\begin{align}
  \begin{split}
    \texttt{simulated\_GD} &\sim \texttt{syn\_category} + \texttt{sentN} \\
    &+ \texttt{tokenN} + \texttt{freq}*\texttt{length} \:\:.
    \label{eq:syn_cat}
    \end{split}
\end{align}
}

\noindent
From this regression model, a log-likelihood score for the simulated gaze duration was obtained.
To evaluate the separate effect of \texttt{syn\_category}, $\Delta$LogLik between Eq.~\ref{eq:syn_cat} and a baseline model was calculated.
The baseline model was $\texttt{simulated\_GD} \sim \texttt{sentN} + \texttt{tokenN} + \texttt{freq}*\texttt{length}$.
The $\Delta$LogLik is denoted as ``Effect of syntactic category.''
A lower score means that the LM lacked the property of varying processing difficulty with respect to the syntactic category.

\paragraph{Results:}
The results are shown in Figure~\ref{fig:syn_ana}.
First, the higher psychometric predictive power the LMs exhibit, the greater the effect of syntactic category on surprisals (left part in Figure~\ref{fig:syn_ana}).
This means that, depending on the syntactic category of the segment they processed, LMs with high psychometric predictive power computed surprisals with a more nonuniform trend.
The right part of Figure~\ref{fig:syn_ana} shows that, as PPL decreases below a certain value ($\mathrm{PPL}\sim400$), the Japanese LMs compute surprisals that obscure the nonuniform trends with respect to the syntactic category of segments.\footnote{The correlation between PPL and the effect of syntactic category in the LMs with PPL less than 400 was 0.45 and 0.34 with Pearson's $r$ and Spearman's $\rho$, respectively.}
This trend supports our hypothesis that tuning to LM objectives obscures the human-like nonuniformity of the processing difficulty.
Even though LMs that are not fully tuned to the LM objective ($\mathrm{PPL}\sim400$) acquire human-like trends with respect to syntactic category, these biases tend to be lost by further lowering their PPL.

Notably, we also observed that not all the types of linguistic nonuniformity were obscured in surprisals computed by the LMs with low PPL.
For example, Appendix~\ref{app:anti-locality} shows that LMs with lower PPL compute surprisals that better correlates with a particular syntactic factor although that factor is a less dominant trend in human reading behavior than the syntactic category  (Appendix~\ref{app:pre_exp2}).

\section{Limitations and future works}
\label{sec:limitations}

To test the universality of the recent findings in computational psycholinguistics across languages, the initial focus is on English and Japanese as a pair of languages with different linguistic properties.
Although the discrepancy of the results in the two languages is discussed from the viewpoint of the UID hypothesis, the two languages are also different in various ways, such as writing systems, agglutinative property, case marking, sentence structure, and pro-drop nature.
To identify the difference that relates to the human-like behaviors of LMs, experiments that include additional languages should be conducted in the future.

In addition, the corpus size of the BCCWJ-EyeTrack data is smaller than the Dundee Corpus. %
While the reading time data in the BCCWJ-EyeTrack was collected from various human subjects, the number of the independent segments was limited (1,643 segments, 218 sentences).
Thus, whether the trends reported in this study generalize to more diverse Japanese texts should be explored in future work.
It is hoped that this study motivates the creation of a large-scale corpus of human reading behaviors in diverse languages.

\section{Conclusion}
\label{sec:con}
This study has investigated whether the recent reports on the psychometric predictive power of LMs can be generalized across languages.
Our initial investigation has re-examined the recent report---\textit{the lower PPL a LM has, the more human-like the LM is}---using Japanese eye movement data.
Our experiments have demonstrated a surprising lack of universality of this report; lower perplexity is not always human-like.
This discrepancy of the results between the languages reinforces the need for the cross-lingual evaluation of the psychometric predictive power of LMs.
The discussion considers potential factors that make the observation different across languages from the viewpoint of the uniform information density hypothesis.
We believe that this is an important first step for seeking a language-agnostic model of human sentence processing.
Hopefully, this study encourages researchers to further investigate the universality of human language processing across languages.

\section*{Acknowledgements}
We would like to thank the members at the Tohoku NLP Lab for their valuable advice, particularly Ana Brassard for proofreading. This work was supported by Grant-in-Aid for JSPS Fellows Grant Number JP20J22697, JSPS KAKENHI Grant Number 19H04990, and JST CREST Grant Number JPMJCR20D2. This work was also supported by the National Institute for Japanese Language and Linguistics (NINJAL) Collaborative Research Project ``Computational Psycholinguistics of Language Processing with Large Corpora.''

\section*{Ethical considerations}
Language models with low perplexity are typically trained with a high computational cost. Our work demonstrates that further up-scaling the models might not be a reasonable direction of searching for human-like language models. This could potentially contribute to reducing energy and carbon costs, which are needed to train large-scale language models.

\bibliography{Mendeley}
\bibliographystyle{acl_natbib}

\clearpage

\appendix

{\setlength\textfloatsep{0pt}
\begin{figure}[t]
    \centering
      \includegraphics[width=7.5cm]{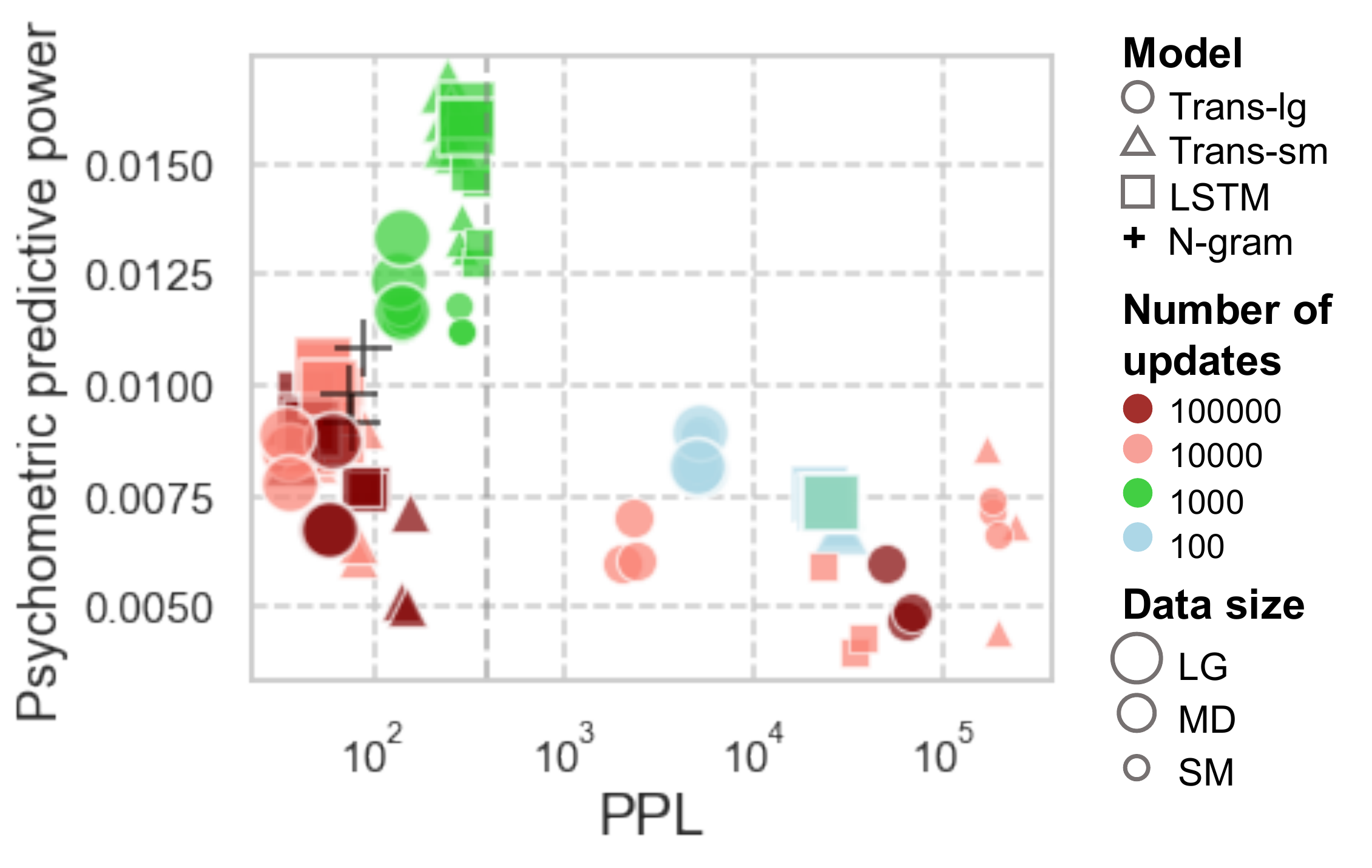}
      \caption{Relationship between PPL (X-axis) and psychometric predictive power (Y-axis). Each point corresponds to each LM. Low score on X-axis indicates the high linguistic accuracy of the model. High score on Y-axis indicates that the model has a high psychometric predictive power. Note that X-axis is on a log scale. The shape, color, and size of each point is same as Figure~\ref{fig:pred_power}.}
      \label{fig:bccwj_fit_logtime}
\end{figure}
}

{\setlength\textfloatsep{0pt}
\begin{figure}[t]
    \centering
      \includegraphics[width=7.2cm]{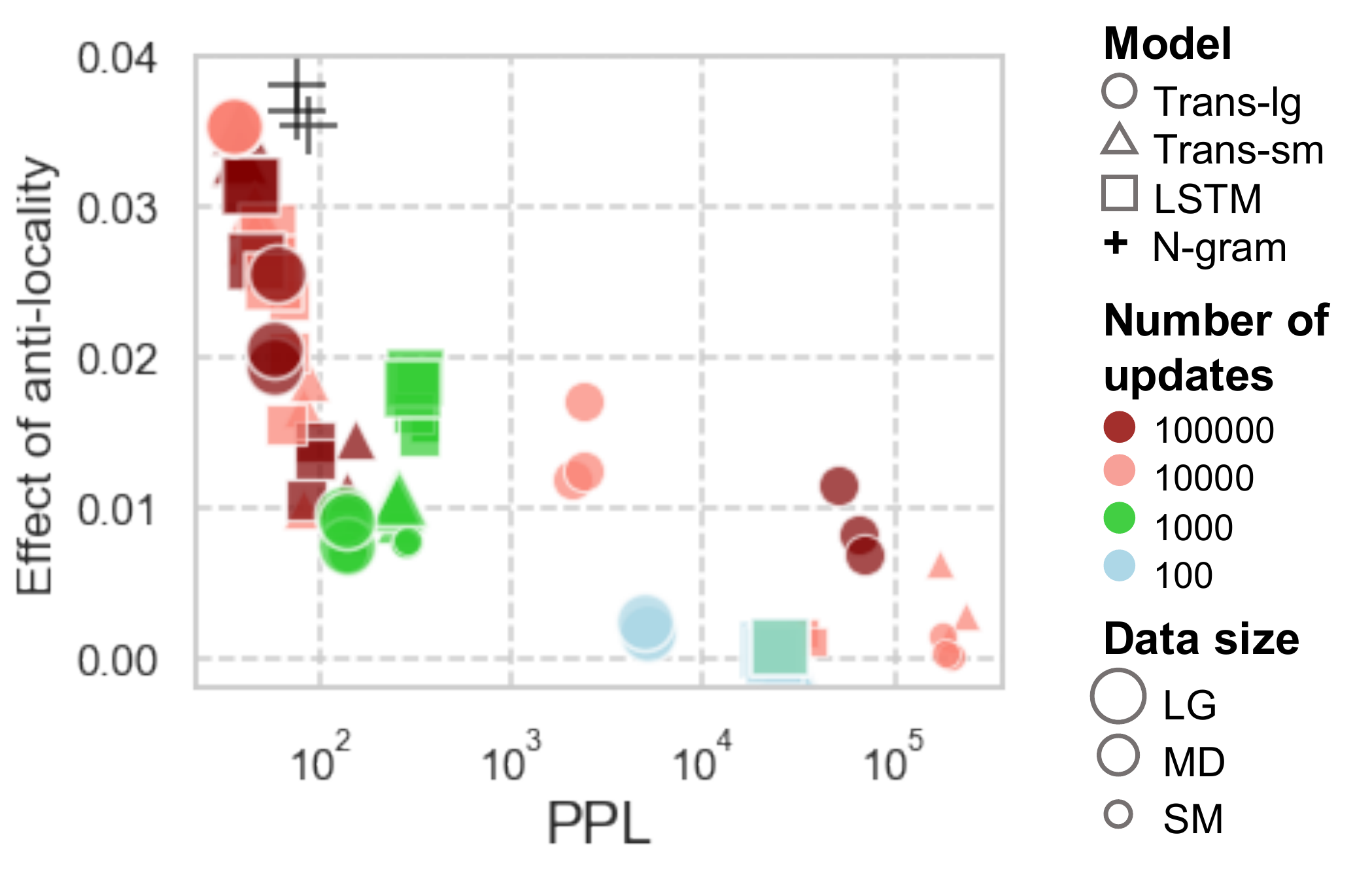}
      \caption{Relationship between PPL (X-axis) and the effect of the anti-locality (Y-axis). Each point corresponds to each LM. A low score on X-axis indicates the high linguistic accuracy of the model. A high score on Y-axis indicates that the surprisals computed by the corresponding model are highly biased towards the anti-locality effect. Note that X-axis is on a log scale. The shape, color, and size of each point is the same as Figure~\ref{fig:pred_power}.}
      \label{fig:syn_str_ana}
\end{figure}
}

\section{Hyperparameters of LMs}
\label{app:hyper}
Table~\ref{tbl:hyper_params} shows the hyperparameters of \textsc{Trans-sm}, \textsc{Trans-lg}, and \textsc{LSTM}, respectively.
Note that the number of parameter updates varies as described in Section~\ref{sec:method}.

\begin{table*}[t]
    \centering
\input{tables/hyper_params}
\caption{Hyperparameters for the LMs.}
\label{tbl:hyper_params}
\end{table*}

\section{Factors used in regression models}
\label{app:feature}
Descriptions for the factors used in our experiments are shown in Table~\ref{tbl:feature_name}.
The frequency of a segment (\texttt{freq}) was estimated using the full training data for the LMs.

\begingroup
\begin{table*}[t]
\input{tables/features.tex}

    \caption{Factor names and their description.}
    \label{tbl:feature_name}
\end{table*}
\endgroup

\begin{table*}[t]
\centering
\begin{tabular}{rrr} \toprule
syntactic category & number of segments & Avg. gaze duration \\
\cmidrule(lr){1-1} \cmidrule(lr){2-2} \cmidrule(lr){3-3}
\texttt{nominal} & 4,322 & 388.4\\
\texttt{verbal} & 1,090 & 291.0 \\
\texttt{modifier} & 588 & 297.1 \\
\texttt{other} & 9 & 239.3 \\
\bottomrule
\end{tabular}
\caption{The statistics of the syntactic category labels in BCCWJ-EyeTrack.}
\label{tbl:syntactic_category}
\end{table*}

\section{Results of modeling logarithmic gaze duration in BCCWJ-EyeTrack}
\label{app:logtime}

Existing studies~\cite{Asahara2016Reading-TimeJapanese} performed experiments using the logarithmic gaze duration because the logarithmic gaze duration more matches the normal distribution than the raw gaze duration.
Given this, we additionally conducted experiments in Section~\ref{sec:exp1}, changing the target variable from the raw gaze duration to its logarithmic gaze duration.
The result with this setting is shown in Figure~\ref{fig:bccwj_fit_logtime}.
There was no substantial difference with the results shown in Section~\ref{sec:exp1}.

\section{Preliminary experiments in Section~\ref{sec:exp2}}
\label{app:pre_exp2}
Which linguistic factor is helpful for explaining the difference in gaze duration?
We conducted experiments using linguistic annotation in the BCCWJ-EyeTrack.
Following the existing studies, we checked the separate effect of syntactic category, semantic category~\cite{Asahara2017BetweenCategories}, and a particular aspect of hierarchical syntactic structure (i.e., the anti-locality effect)~\cite{Asahara2016Reading-TimeJapanese}.
Specifically, we used the factors, \texttt{syn\_category}, \texttt{sem\_category}, and \texttt{n\_dependents}, shown in Table~\ref{tbl:feature_name}.
For each factor, we inspect the separate effect of each factor for modeling gaze duration.
As Eq.~\ref{eq:syn_cat}, we first modeled the gaze duration using each factor (\texttt{factor\_X}):

{\small
\begin{align}
\begin{split}
    \texttt{GD} &\sim \texttt{factor\_X} + \texttt{sentN} \\
    &+ \texttt{segmentN} + \texttt{freq}*\texttt{length} \:\:.
    \label{eq:annotation}
\end{split}
\end{align}
}

\noindent
Then, we calculated the $\Delta$LogLik between X and a baseline model.
The baseline model was $\texttt{GD} \sim \texttt{sentN} + \texttt{segmentN} + \texttt{freq}*\texttt{length}$.

The $\Delta$LogLik for each collection of factors are shown in~\ref{tbl:dominance}.
We found that syntactic category is the most influential factor for modeling gaze duration, at least in this experiment.

\begin{table*}[t]
\centering
\begin{tabular}{rr} \toprule
linguistic property & $\Delta$LogLik \\
\cmidrule(lr){1-1} \cmidrule(lr){2-2}
syntactic category & 58.37 \\
semantic category & 17.08 \\
number of dependents & 13.84 \\
\bottomrule
\end{tabular}
\caption{The separate effect of each linguistic annotation for modeling gaze duration.}
\label{tbl:dominance}
\end{table*}

\section{Anti-locality effect in LMs}
\label{app:anti-locality}

Similar to Section~\ref{sec:exp2}, we analyzed how strongly the surprisals from each Japanese LM are biased towards a particular linguistic property.
In this section, we investigated the anti-locality effect in the surprisals from LMs.
The anti-locality is that the more dependents a segment has in its preceding context, the cognitive effort of the head segment is reduced  (i.e., modifiers alleviate the processing cost of their head).

Analogous to the Section~\ref{sec:exp2}, we regarded surprisal as ``simulated gaze duration'' from an ``LM subject,'' and evaluated the importance of the number of the dependents in its preceding context (\texttt{n\_dependents}) for modeling the simulated gaze duration (\texttt{simulated\_GD}).
To inspect the effect of the \texttt{n\_dependents} for modeling the simulated gaze duration, we used the following regression model:

{\small
\begin{align}
\begin{split}
    \texttt{simulated\_GD} &\sim \texttt{n\_dependents} + \texttt{sentN} \\
    &+ \texttt{tokenN} + \texttt{freq}*\texttt{length} \:\:.
    \label{eq:syn_str}
\end{split}
\end{align}
}

\noindent
From this regression model, we obtained a log-likelihood score for the simulated gaze duration.
To evaluate the separate effect of \texttt{n\_dependents}, we calculated the $\Delta$LogLik between Eq.~\ref{eq:syn_str} and a baseline model.
The baseline model was $\texttt{simulated\_GD} \sim \texttt{sentN} + \texttt{segmentN} + \texttt{freq}*\texttt{length}$.
The $\Delta$LogLik is denoted as ``Effect of the anti-locality.''

The results are shown in Figure~\ref{fig:syn_str_ana}.
There is a clear trend that the LMs with lower PPL exhibit surprisals that are more consistent with the anti-locality effect (Spearman's $\rho=-0.77$ between PPL and the strength of the anti-locality effect).
This suggests that the surprisals from LMs with low PPL are biased towards the hierarchical structure of sentences rather than the syntactic category.

\end{document}

%% file: tables/hyper_params.tex
\begin{minipage}[t]{\hsize}
\renewcommand{\arraystretch}{0.9}
    \centering
    {\small
    \begin{tabular}{llc} \toprule
     \multirow{9}{*}{Fairseq model} & architecture & transformer\_lm\_gpt2\_small \\
      & adaptive softmax cut off & 50,000, 140,000 \\
      & share-decoder-input-output-embed & True \\
      & embed\_dim & 1,024 \\
      & ffn\_embed\_dim & 4,096 \\
      & layers & 24 \\
      & heads & 16 \\
      & dropout & 0.1 \\
      & attention\_dropout & 0.1 \\
    \cmidrule(lr){1-1} \cmidrule(lr){2-2} \cmidrule(lr){3-3}
    \multirow{5}{*}{Optimizer} & algorithm & AdamW \\
    & learning rates & 5e-4 \\
    & betas & (0.9, 0.98) \\
    & weight decay & 0.01 \\
    & clip norm & 0.0 \\
    \cmidrule(lr){1-1} \cmidrule(lr){2-2} \cmidrule(lr){3-3}
    \multirow{3}{*}{Learning rate scheduler} & type & inverse\_sqrt \\
    & warmup updates & 4,000 \\
    & warmup init lrarning rate & 1e-7 \\
    \cmidrule(lr){1-1} \cmidrule(lr){2-2} \cmidrule(lr){3-3}
    \multirow{2}{*}{Training} & batch size & 61,440 tokens \\
    & sample-break-mode & none \\ \bottomrule
        \end{tabular}
        }
        \subcaption{\textsc{Trans-lg}.}
        \label{tbl:hyperparam_tl}
        \vspace{0.2cm}
\end{minipage}

\begin{minipage}[t]{\hsize}
\renewcommand{\arraystretch}{0.9}
    \centering
    {\small
    \begin{tabular}{llc} \toprule
     \multirow{9}{*}{Fairseq model} & architecture & transformer\_lm\_gpt \\
      & adaptive softmax cut off & 50,000, 140,000 \\
      & share-decoder-input-output-embed & True \\
      & embed\_dim & 384 \\
      & ffn\_embed\_dim & 2,048 \\
      & layers & 8 \\
      & heads & 6 \\
      & dropout & 0.1 \\
      & attention\_dropout & 0.1 \\
    \cmidrule(lr){1-1} \cmidrule(lr){2-2} \cmidrule(lr){3-3}
    \multirow{5}{*}{Optimizer} & algorithm & AdamW \\
    & learning rates & 5e-4 \\
    & betas & (0.9, 0.98) \\
    & weight decay & 0.01 \\
    & clip norm & 0.0 \\
    \cmidrule(lr){1-1} \cmidrule(lr){2-2} \cmidrule(lr){3-3}
    \multirow{3}{*}{Learning rate scheduler} & type & inverse\_sqrt \\
    & warmup updates & 4,000 \\
    & warmup init lrarning rate & 1e-7 \\
    \cmidrule(lr){1-1} \cmidrule(lr){2-2} \cmidrule(lr){3-3}
    \multirow{2}{*}{Training} & batch size & 61,440 tokens \\
    & sample-break-mode & none \\ \bottomrule
        \end{tabular}
        }
        \subcaption{\textsc{Trans-sm}.}
        \label{tbl:hyperparam_ts}
        \vspace{0.2cm}
\end{minipage}

\begin{minipage}[t]{\hsize}
\renewcommand{\arraystretch}{0.9}
    \centering
    {\small
    \begin{tabular}{llc} \toprule
     \multirow{7}{*}{Fairseq model} & architecture & lstm\_lm \\
      & adaptive softmax cut off & 50,000, 140,000 \\
      & share-decoder-input-output-embed & True \\
      & embed\_dim & 400 \\
      & hiden\_size & 1,024 \\
      & layers & 2 \\
      & dropout & 0.1 \\
    \cmidrule(lr){1-1} \cmidrule(lr){2-2} \cmidrule(lr){3-3}
    \multirow{5}{*}{Optimizer} & algorithm & AdamW \\
    & learning rates & 1e-3 \\
    & betas & (0.9, 0.98) \\
    & weight decay & 0.01 \\
    & clip norm & 0.0 \\
    \cmidrule(lr){1-1} \cmidrule(lr){2-2} \cmidrule(lr){3-3}
    \multirow{3}{*}{Learning rate scheduler} & type & inverse\_sqrt \\
    & warmup updates & 4,000 \\
    & warmup init lrarning rate & 1e-7 \\
    \cmidrule(lr){1-1} \cmidrule(lr){2-2} \cmidrule(lr){3-3}
    \multirow{2}{*}{Training} & batch size & 20,480 tokens \\
    & sample-break-mode & none \\ \bottomrule
        \end{tabular}
        }
        \subcaption{\textsc{LSTM}.}
        \label{tbl:hyperparam_lstm}
        \vspace{0.2cm}
\end{minipage}

%% file: tables/features.tex
\centering
\begin{tabular}{llp{10cm}} \toprule
Factor name & Type & Description \\ 
\cmidrule(lr){1-1} \cmidrule(lr){2-2} \cmidrule(lr){3-3}
\texttt{surprisal} & num & surprisal caluzulted by LMs \\
\texttt{GD} & num & reading time (first pass time)\\
\texttt{article} & factor & article ID\\
\texttt{screenN} & int & screen display order \\
\texttt{lineN}& int & the serial number of line the segment is displayed \\
\texttt{segmentN} & int & the serial number of segment in a screen \\
\texttt{sentN} & int & the serial number of sentence the segment belongs to\\
\texttt{tokenN} & int & the position of segment in sentence \\
\texttt{length} & int & number of characters \\
\texttt{freq} & num & geometric mean of the frequencies of subword constituents in a segment \\
\texttt{subj} & factor & participant ID \\
\cmidrule(lr){1-1} \cmidrule(lr){2-2} \cmidrule(lr){3-3}
\texttt{syn\_category} & factor & syntactic category the segment falls into (\texttt{nominal}, \texttt{verbal}, \texttt{modifier}, or \texttt{other}) \\
\texttt{sem\_category} & factor & semantic category the segment falls into (\texttt{relation}, \texttt{subject}, \texttt{action}, \texttt{product}, or \texttt{nature}) \\
\texttt{n\_dependents} & int & number of dependents before the segment \\
\bottomrule
\end{tabular}